# Evolving Deep Neural Network by Customized Moth Flame Optimization Algorithm for Underwater Targets Recognition

*Mohammad Khishe[1] [*], Mokhtar Mohammadi[2], Tarik A. Rashid[3], Hoger Mahmud [4], Seyedali Mirjalili[5,6]*

[1*] *Corresponding author, Department of Marine Electronics and Communication Engineering, Imam Khomeini Marine Science University, Nowshahr, Iran,* m_khishe@iust.alumni.ac.ir

[2] *Department of Information Technology, College of Engineering and Computer Science, Lebanese French University, KR-Iraq.*

[3] *Computer Science and Engineering Department, Science and Engineering Science, University of Kurdistan Hewler, Erbil, KR-Iraq.*

[4] *Department of Computer Science, College of Science and Technology, University of Human Development, Sulaymaniyah, KR-Iraq.*

[5] *Centre for Artificial Intelligence Research and Optimization, Torrens University, Australia,*

[6] *Yonsei Frontier Lab, Yonsei University, Seoul, Korea.*

**Abstract:** This chapter proposes using the Moth Flame Optimization (MFO) algorithm for fine-tuning a Deep Neural Network to recognize different underwater sonar datasets. Same as other models evolved by metaheuristic algorithms, premature convergence, trapping in local minima, and failure to converge in a reasonable time are three defects MFO confronts in solving problems with high-dimension search space. Spiral flying is the key component of the MFO as it determines how the moths adjust their positions in relation to flames; thereby, the shape of spiral motions can regulate the transition behavior between the exploration and exploitation phases. Therefore, this chapter investigates the efficiency of seven spiral motions with different curvatures and slopes in the performance of the MFO, especially for underwater target classification tasks. To assess the performance of the customized model, in addition to benchmark Sejnowski & Gorman's dataset, two experimental sonar datasets, i.e., the passive sonar and active datasets, are exploited. The results of MFO and its modifications are compared to four novel nature-inspired algorithms, including Heap-Based Optimizer (HBO), Chimp Optimization Algorithm (ChOA), Ant Lion Optimization (ALO), Stochastic Fractals Search (SFS), as well as the classic Particle Swarm Optimization (PSO). The results confirm that the customized MFO shows better performance than the other state-of-the-art models so that the classification rates are increased 1.5979, 0.9985, and 2.0879 for Sejnowski & Gorman, passive, and active datasets, respectively. The results also approve that time complexity is not significantly increased by using different spiral motions.

## 1. Introduction

Underwater target classification is a complex and challenging process that has recently been the subject of many studies [1–3]. A practical classification task requires careful consideration for mechanisms that can differentiate between target and non-target object echoes as well as the echoes of objects that lurk in the background; classifications can also be affected by dependencies such as operation mode and engine procedures. Moreover, according to [3], the phase and amplitude of the acoustic signal received from objects can be altered by signal characteristics such as the temporal variation of propagation channel, dispersion, and ambient noises. To address the challenge, researchers have proposed two main classification methods, which are deterministic and stochastic. The first method utilizes statistical processing, oceanography, and models generated by sonar technologies to perform the classification [4]. The second method makes use of artificial intelligence to extract underwater features and predict and estimate underwater phenomena [5]. Deterministic classification calculates the statistical distribution of several parameters obtained from sources such as environmental conditions, sound propagations, topographic maps, and background clutters [6]. However, the authors of [7] have stated several disadvantages that may affect the method's applicability, such as the high cost of required equipment and human resource and lack of guarantee in terms of robustness and generality. The stochastic approach focuses on cost and complexity reduction in sonar target classification using Intelligent Machine Hearing (IMH) to locate and identify acoustic signals coming from underwater objects [8].

Despite the excellent accuracy, the IMH-based methods are tedious because they first divide a problem into several sub-sections and then accumulate the result. Besides, IMH cannot handle big datasets, resulting in low accuracy. Moreover, canonical IMHs may not be able to learn cluttered backscattered so that they rely more on the objects' features than employing the whole background data [9].

These shortcomings can be addressed using Deep Learning (DL) approaches, which are end-to-end models efficiently working with enormous data [10]. However, although DL's outstanding features enable it to resolve different learning issues, training is challenging [11]. By reviewing the literature, it can be realized that eminent algorithms for training DL are Gradient-descendent [12], conjugate gradient algorithm [13], Hessian-free optimization approach [14], and Krylov subspace descent algorithm [15].

Although stochastic gradient descent-based training algorithms have simple structures and fast convergence rates for large training samples, gradient descent-based approaches need numerous manual parameter tuning for optimality. Their design is also sequential, which makes parallelizing with Graphics Processing Units (GPUs) challenging. Simultaneously, though conjugate gradient methods are stable for training, they are almost slow, demanding multiple CPUs and many memory resources [13]. Deep auto-encoders utilized Hessian-free optimization for training the weights of typical Convolutional Neural Network (CNN) [14], which is more effective in pre-training and fine-tuning deep auto-encoders than the model introduced by Hinton and Salakhutdinov [16]. On another note, the Krylov subspace descent algorithm is more stable and more straightforward than Hessian-free optimization. Thereby, it presents a more accurate classification rate and optimization speed rather than Hessian-free optimization. On the contrary, the Krylov subspace descent algorithm needs more memory than Hessian-free optimization [16].

Recently, evolutionary and metaheuristic algorithms have been widely applied to optimize different engineering optimization tasks [17]. Nevertheless, research studies on the metaheuristic algorithm for optimizing DL models require more studies. The hybrid Genetic Algorithm (GA) and CNN, introduced in [18], was the first study that initiates this kind of optimization model using metaheuristic optimization algorithms. This model selects the CNN parameters utilizing the GA's cross-over and mutation operation, in which the structure of DCNN is modeled as a GA's chromosome. Besides, in the cross-over phase, the first convolution layer's biases and weights (C1) and the third convolution layer (C3) are utilized as chromosomes. Reference [19] proposed an evolutionary fine-tuning method to optimize the parameters of a DCNN using the Harmony Search (HS) algorithm and some of its improved versions in the field of handwritten digit and fingerprint recognition. To avoid local minima and improve performance, the authors of [20] combined back-propagation with GSA and PSO algorithms to develop a new CNN algorithm. To improve canonical PSO's exploration capability, the authors of [21] proposed cPSO-CNN, a variant of PSO, to enhance architecture-determined CNNs in terms of hyper-parameter configuration. Their reported results show that the cPSO-CNN enhances hyper-parameter configuration and reduces overall computation costs compared to several proposed algorithms in the literature.

The hybrid tabu-genetic algorithm [22], grey wolf optimizer [23], simulated annealing and differential evolution algorithms [24], micro-canonical annealing algorithm [25], improved cuckoo search algorithm [26], whale optimization algorithm [27], etc. are other novel metaheuristic and nature-inspired algorithms, which have used to optimized the CNN's performance. However, according to the theory of No Free Lunch (NFL) [28], a comprehensive optimization algorithm capable of addressing all optimization issues is yet to be developed, leading to the utilization of nature-inspired algorithms to develop novel and more capable algorithms by researchers. Following their directions, this chapter proposes the application of MFO algorithm which is a variant of nature-inspired metaheuristic algorithm to improve DCNN classification performed on datasets obtained from underwater sonar equipment.

MFO algorithm is developed based on navigation paths that moths carve in nature. Researchers' experimental results show that MFO is highly effective for avoiding local minima [29–31] and MLP and RBF training [29,30]. The results also show that MFO produces excellent results in the exploration phase; however, its performance is not equally excellent in the exploitation phase of sonar dataset search spaces. This chapter modifies MFO with spiral motions to optimize the algorithm for sonar dataset classification and DCNNs training.

To evaluate the effectiveness of the modified MFO, three underwater acoustic datasets, including synthetic, Sejnowski & Gorman's dataset [31], passive sonar dataset [32], and active sonar dataset [33], are exploited. To ensure a thorough comparison, the efficiency of the designed model is compared with ten recently proposed benchmark models.

The book chapter is organized as follows: Background materials, including DCNN, sonar datasets, and MFO, will be reviewed in Section 2. Section 3 introduces the customized MFO and its fine-tuning scheme. Section 4 represents the simulation results and discussion, and conclusions are finally described in Section 5.

## 2. Methods

This section provides background information, such as the DCCN model, underwater sonar datasets, and MFO.

### 2.1. Deep neural network architectures

Neurons connect and collaborate to accomplish intelligence functionalities known as Neural Network (NN). Deep NN is a form of NN which is classified into discriminative and generative categories. The discriminative category utilizes a bottom-up approach to generate labels for hidden classes of data using maximum calculated similarities amongst them. The second category utilizes a top-down method to create a model for hidden data classes based on the distribution of observable classes.

Deep NN may be classified further into four general classes based on architectural properties as presented in Fig. 1: Unsupervised Pretrained Networks (UPN) [34], Convolutional Neural Networks (CNNs) [35], Recurrent Neural Networks (RNNs) [36], and Recursive Neural Networks. UPNs include Auto-Encoders (AEs) [37], Restricted Boltzmann Machines (RBMs) [38], Generative Adversarial Networks (GANs) [39], and Deep Belief Networks (DBNs) [40]. Long Short-Term Memory (LSTM) [41] is also an implementation of RNNs.

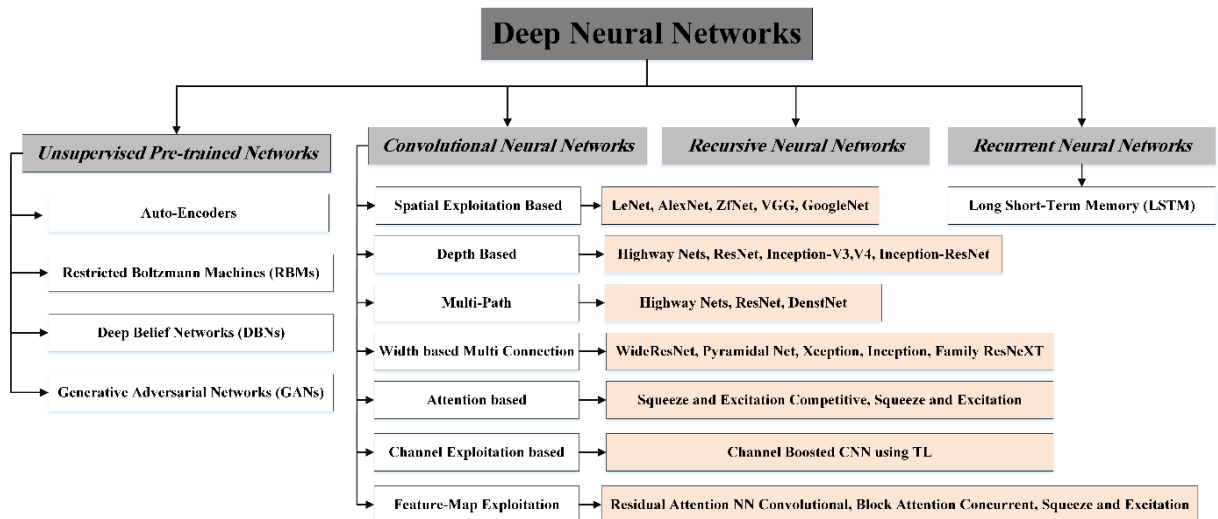

**Fig. 1:** A general block diagram for DNN architectures.

### 2.2. Convolutional neural networks

One of the most widely used DNN models in studies related to data structures in grids such as images is CNN. The model consists of several layers: dense, dropout, reshape, flatten, convolution, and pooling. In the model, calculation and transformation tasks are performed independently and passed on to the next layer. The dimension of the hidden layer specifies the model's width, and the depth of the layer determines the length of the model. Fig. 2 shows the feature map of a CNN model (LeNet-5 DCNN) comprised of two convolution and pooling layers [42]. Each CNN layer performs some specific tasks; for instance, the convolution layer uses probability density function to weight elements in a data matrix singly. Layers are connected through neurons; for instance, convolution layer neurons are connected to the receptive points of the previous layer; any neurons with matching Feature Maps (FMs) can receive data from multiple input neurons until the input is wholly swiped to share identical connection weights. The pooling layers down-sample the FMs by a factor of 2. For example, in layer ($C_3$), the FM of size $12 \times 12$ is sub-sampled to the FM of size $6 \times 6$ in the following layer ($S_4$). The structure presented in Fig. 2 linear filter and kernels are used to calculate each FM, which results from a convolution from the prior layer's maps. The adding bias $b_k$ and weights $w^k$ produce the $FM_{ij}^k$ using the *tanh* function as shown in Eq. (1).

$$FM_{ij}^k = \tanh((W^k \times x)_{ij} + b_k) \qquad (1)$$

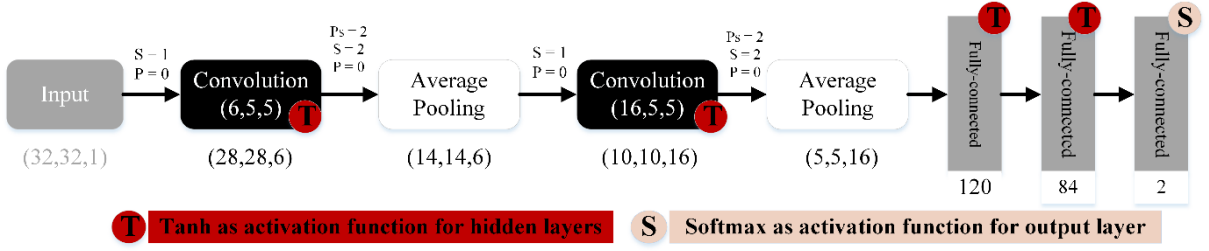

**Fig. 2:** The LeNet-5 DCNN

Sub-sampling results in spatial invariance when FMs resolution is reduced, and each FM refers to the prior layer. Eq. (2) defines the pooling model.

$$\alpha_j = \tanh(\beta \sum_{N \times N} \alpha_i^{n \times n} + b) \qquad (2)$$

Where $\beta$ is a trainable scalar, $\alpha_i^{n \times n}$ are the inputs, and $b$ is a bias. The last layer, which is the fully connected layer, performs the classification and comes after several convolution and sub-sampling layers. It should be noted that for each output class, there is one neuron.

**2.3. Sonar Dataset**

For the aim of having a complete inquiry, the designed classifiers are tested against three underwater sonar datasets as follows:

- **Sejnowski & Gorman's Underwater Sonar Dataset** [31]**:**
  This freely available dataset is produced in Sejnowski & Gorman's experiment [31], which is used in this chapter to evaluate the suggested classifier in comparison to other classifiers presented in the literature. Their experiment collected data regarding two echoes coming from a 5 feet metal cylinder and a rock with the same property. The metal cylinder served the purpose of a real target, and the rock served the purpose of the false target. In order to collect the data, a wideband linear modulated chirp ping is aimed towards the objects, which are positioned on a sandy bottom. During the experiment, Sejnowski & Gorman's collected 1200 echoes from which 208 were selected, out of which 111 echoes belong to the metal cylinder, and 97 are related to the same sized rock. In selecting the 208 echoes, 4-15 dB Signal to Noise Ratio (SNR) is used as the threshold level.

- **Passive Sonar Dataset** [32]**:**
  The authors of [32] generated a new passive underwater sonar dataset from their experiment in which they used different water velocities and densities to test seven different types of propellers at various Revolutions Per Minute (RPM). The propellers were laboratory samples that were the same as the originals, and the experiment was carried out in a testing tunnel model B&K_8103. Fig. 3 shows a typical propeller noise sample, Fourier transform, hydrophone noise level, and Power Spectral Density (PSD) estimate. In order to test the proposed classifier in this chapter, the spectrogram of the input signal is required. Fig.4 shows a sample of a typical spectrogram prior and post noise removal phase.

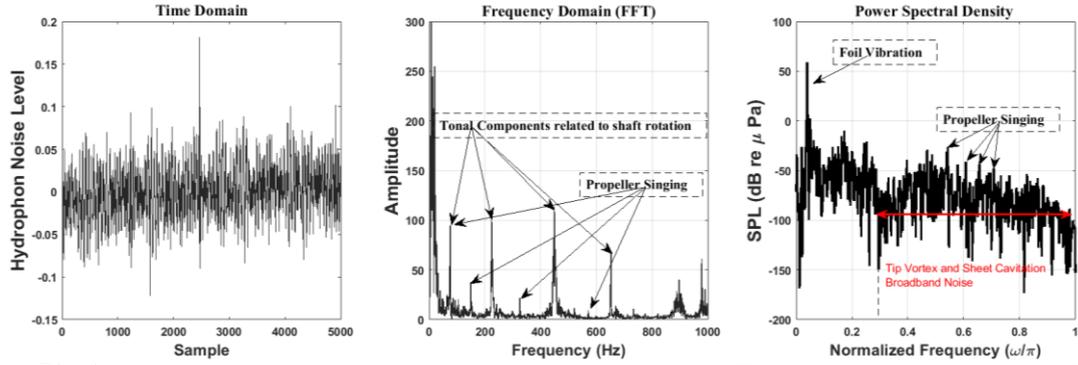

**Fig.3:** A typical hydrophone received propeller sound, Fourier transform, and its PSD

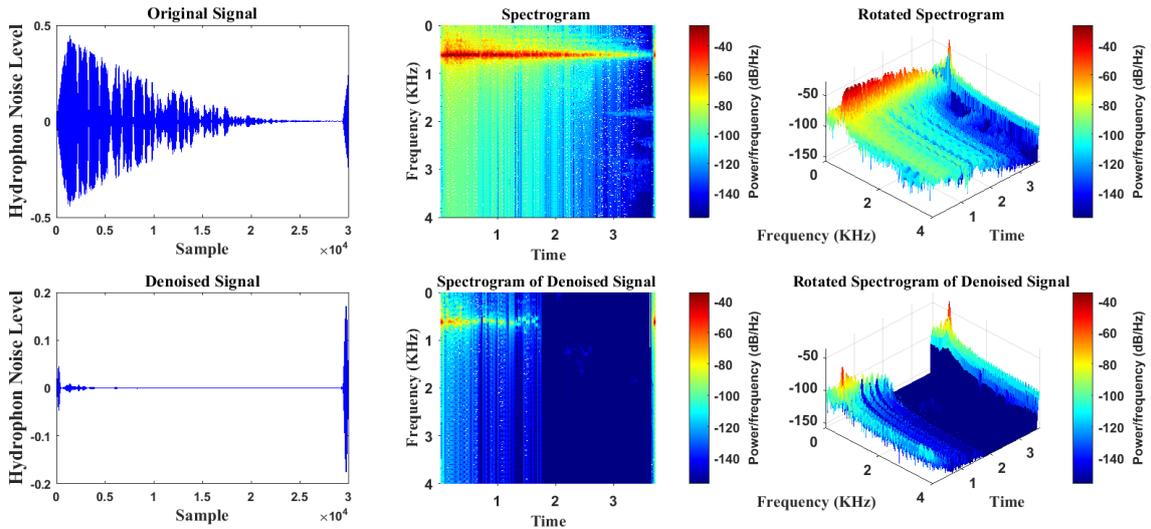

**Fig.4:** A typical spectrogram of radiated noise and its denoised spectrogram.

- **Active Sonar Dataset** [33]**:**

This dataset is generated by Khishe et al. in an experiment they carried out for four targets and two non-target objects placed in a sandy seabed. They used a multi-task sonobuoy owned by the Port and Maritime Organization to collect the data from the objects. In order to collect the data, they rotated the objects 180 degrees with one-degree precision using an electric motor. The transmitted signal towards the objects covered a frequency range of 5-110Hz and used a wideband linear frequency modulated ping signal. The process requires complex calculations as the amount of raw data obtained is large, and to lessen the complexity, the authors have relied on a detection process presented in [43]. A sample of typical echoes received from the target and non-target objects is shown in Fig. 5. In [43], the authors have also described a feature extraction process to put together the dataset, such as the one presented in Table 1, which includes the number of samples, training and test instances, and classes.

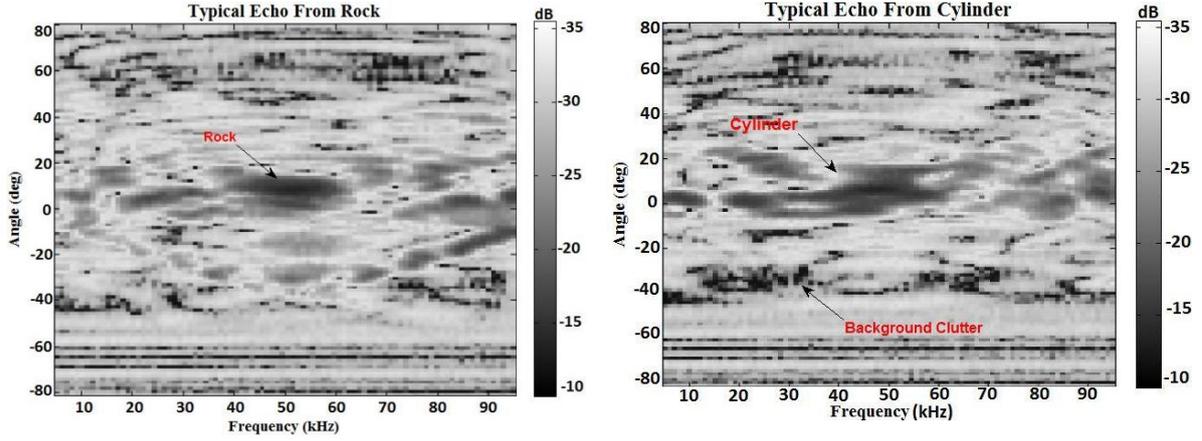

a) Non-target                              b) Target

**Fig.5**: Typical received echoes belong to a target and a non-target.

Table 1: the description of the utilized datasets.

| Datasets | Sejnowski & Gorman | Active Sonar Dataset | Passive Sonar Dataset |
|---|---|---|---|
| No. Training Samples | 150 | 200 | 400 |
| (After data augmentation) | (1200) | (1600) | (3200) |
| No. Test Samples | 58 | 150 | 250 |
| (After data augmentation) | (464) | (1200) | (2000) |
| No. Classes | 2 | 6 | 7 |

It is worth stating that Sejnowski & Gorman sonar dataset is selected to be used as a benchmark to compare the proposed algorithm with others in the literature, while the passive and active datasets presented in [32-34] are selected to be used as a real-world practical experiment.

### 2.4. Moth-Flame Optimization Algorithm

MFO, a population-based nature-inspired algorithm introduced by Seyedali Mirjalili in [44]. It is inspired by the way moths move at night, in which they keep a constant angle with the moonlight in order to stay on a straight path; this approach is known as traversing orientation for navigation. MFO is mathematically modeled using the movement of moths in nature when subjected to artificial light. In such a case, moths converge with the artificial light and proceed in a closed spiral motion, as pictured in Fig. 6.

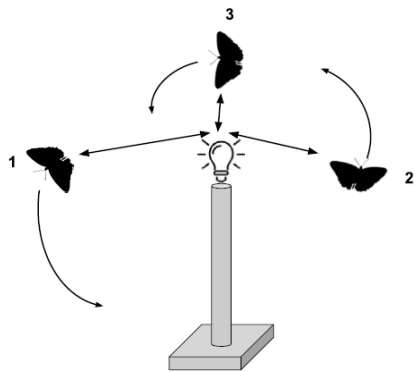

**Fig. 6:** Diagrammatic illustration of moths' spiral movement when exposed to close artificial light source [44].

It is clear by now that moth and flame are the two main components of MFO; the moth represents the search agent, and the position of the moth concerning the light source represents problem variables. Thus the targeted optimization problem's dimension is the length of the position; based on that, the moths' population is expressed in Eq. (3) [44].
6

$$\text{position} = \begin{bmatrix} \text{moth}_{11} & \text{moth}_{12} & \dots & \text{moth}_{1\text{dim}} \\ \text{moth}_{21} & \text{moth}_{22} & \dots & \text{moth}_{2\text{dim}} \\ \vdots & \vdots & \vdots & \vdots \\ \text{moth}_{n1} & \text{moth}_{n2} & \dots & \text{moth}_{n\text{dim}} \end{bmatrix} \quad (3)$$

In the equation, the moths' number in a population is represented by *n*, and the number of variables in a problem is represented by *dim*. The algorithm utilizes a fitness function to investigate each moth's fitness in a population and uses a matrix to store the obtained fitness value. For this research, let us call the matrix Outcome, which can be formulated as Eq. (4) [44], and $Outmoth_n$ represents the fitness value of the moth number *n*.

$$\text{Outcome} = \begin{bmatrix} \text{Outmoth}_1 \\ \text{Outmoth}_2 \\ \vdots \\ \text{Outmoth}_n \end{bmatrix} \quad (4)$$

As flames are the second component, MFO preserves an equal number of flames to that of the number of moths in a population. Flames are also described as a matrix with the same property of Eq. (3). Flames indicate the best position each moth has obtained, and the fitness value for each flame is stored in the same manner as the population *Position* in Eq. (4).

MFO creates a random population of moths through its repeated search process, and moths in a population memorize their position in the form of a flame. The algorithm then performs a fitness evaluation for each moth and stores its fitness value. The moths' positions and their flame values are updated in each repeated search process, and the variable lower and upper bound that represent the solution is checked after each update. The updating process ends when the maximum number of iteration is achieved.

Eq. (5) presented in [44] is used to calculate each moth's position based on a logarithmic spiral. The distance between a moth (represented as $moth_i$) and its flame (represented as $flame_j$) is calculated and stored as $D_{ij}$ using Eq. (6) [44]. In Eq. (5), the logarithmic spiral shape is represented as *q*, and *t* denotes a random number in the range of [*r*, 1].

$$F(moth_i, flame_j) = D_{ij} \cdot e^{qt} \cos(2 \times \pi \times t) + flame_j \quad (5)$$

$$D_{ij} = |flame_j - moth_i| \quad (6)$$

As explained in [44], to preserve the exploitation capability of MFO, Eq. (7) is used to reduce the number of flames (represented as $flmaes_{No}$) as the number of iterations increases. Thus, in the equation, the current iteration number is represented as *l*, and the flames' maximum number and iteration are represented as *n* and *T*, respectively.

$$\text{flames}_{No} = round(n - l \times \frac{n-1}{T}) \quad (7)$$

## 3. Methodology
In the following, the concepts of the customized MFO, the presentation of search agents, and the loss function are described.

### 3.1. Customized MFO
As stated, there is a gap in the literature when it comes to selecting a suitable spiral shape for investigating the aptitude of MFO in the exploitation phase. In this regard, this section investigates several renowned spiral shapes. Generally speaking, a spiral shape is formed when a line is circling outwards around a center point, and they come in different types and shapes. Table 2 presents six renowned types of spirals and their mathematical models presented by the authors of [3,45], and Fig. 7 shows the two and three-dimensional view of each of the six spirals.

Table 2: The utilized spirals and their mathematical models

| Row | Spirals | Equation |
|---|---|---|
| 1 | Archimedean | $r = a \times \theta$ |
| 2 | Logarithmic | $\log(r) = a \times \theta$ |
| 3 | Fermat | $r^2 = a^{2 \times \theta}$ |
| 4 | Lituus | $r^2 = a^2/\theta$ |
| 5 | Equiangular | $\ln(r) = a \times \theta$ |
| 6 | Random | $r = \text{rand}() \times \theta$ |

Where, $r$ represents the radius of spirals, $\theta$ denotes the polar angle, and $a$ is any constant [3,45].

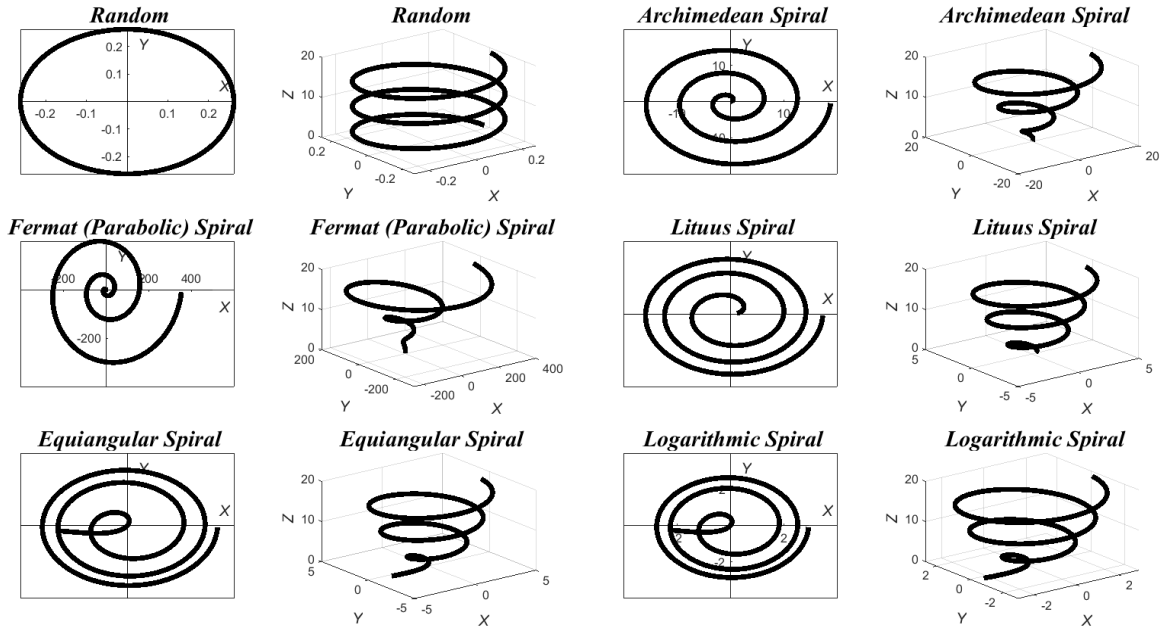

**Fig.7:** The illustrations of various spiral motions.

### 3.2. Presentation of Searching Agents

In order to use an optimization algorithm in the DCNN fine-tuning task, two conditions have to be fulfilled; first, the search agent must represent the parameters of the structure distinctly, and second, the investigated problem must be the base for defining the fitness function. Furthermore, to use MFO in fine-tuning DCNN setup to achieve the maximum detection accuracy, network parameters such as weights and biases must be presented clearly for the fully connected layers.

In such as case, the weight and bias of the last layer are optimized by MFO, and the loss function then calculates the fitness function. Furthermore, MFO uses the bias and connection weight values as moths. In general, binary-based, vector-based, and matrix-based [24] are the schemes used to present a DCNN weight and biases as the MFO possible solution. Since MFO requires the parameters in vector-based form, Eq. (8) is used in this chapter to represent the possible solutions.

$$\text{Moths} = [W_{11}, W_{12}, \ldots, W_{nh}, b_1, \ldots, b_h, M_{11}, \ldots, M_{hm}] \qquad (8)$$

In this equation, $W_{ij}$ denotes the connection weight from the $i_{th}$ input neuron to the $j_{th}$ hidden node, $n$ indicates the input neurons' number, $b_j$ shows the $j_{th}$ bias of hidden neuron, and $M_{jo}$ denotes the connection weight between the $j_{th}$ hidden neuron and the $o_{th}$ output neuron. As previously stated, the proposed architecture is a simple LeNet-5 structure. The convolution layer has a kernel size of *5×5*, and the down-sampling is performed by a factor of two.

### 3.3. Loss function

The main objectives of using MFO to train DCNN are to minimize classification errors and acquire maximum accuracy. The Mean Square Error (MSE) is usually considered as loss function in classification tasks; in this regard, Eq.(9) shows the loss function as the fitness function for DCNN-MFO.

$$y = \frac{1}{2}\sqrt{\frac{\sum_{i=0}^{N}(o-d)^2}{N}} \tag{9}$$

In the equation, the calculated output is represented by *o*, the desired output is represented by *d,* and the training sample is represented by *N*. The MFO uses maximum iteration reached or specified loss function as termination criteria. Fig. (8) shows the flowchart of the designed DCNN-MFO.

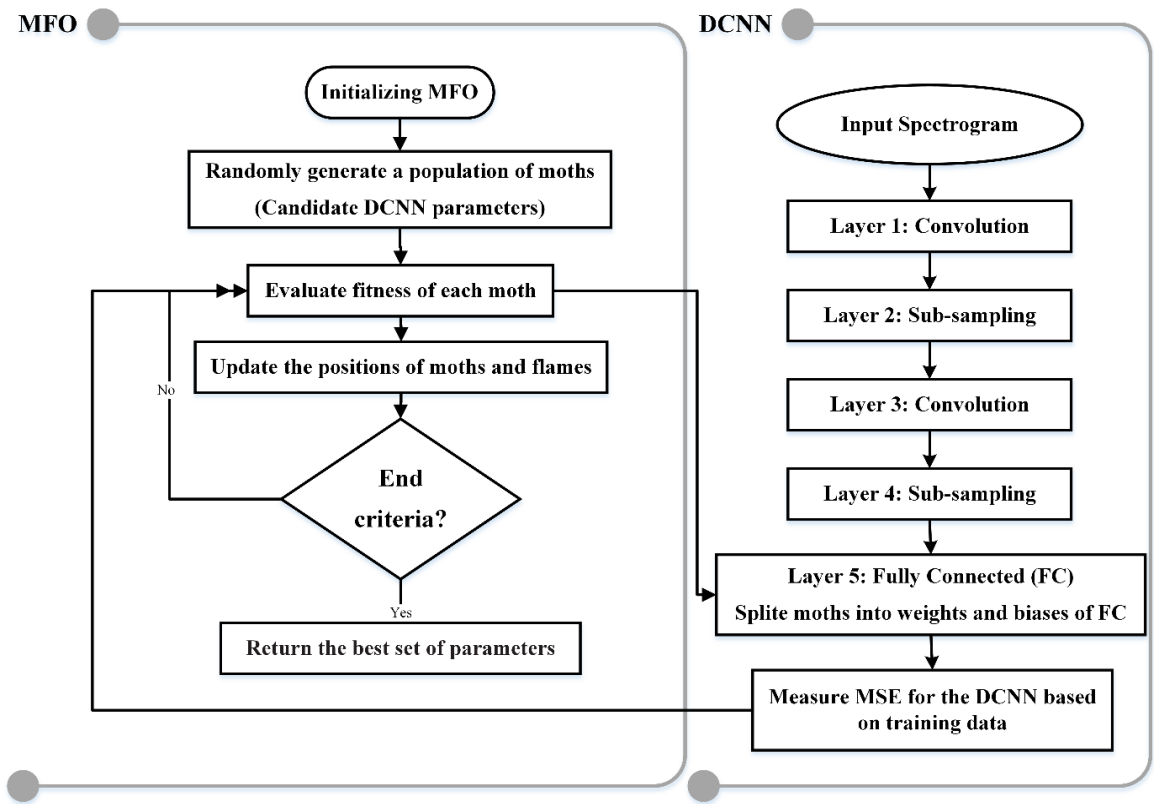

**Fig. 8:** The DCNN-MFO's flowchart.

## 4. Simulation Results and discussion

In order to examine the performance of the proposed DCNN-MFO, the datasets presented in Table 1 are used and grouped as Sejnowski & Gorman, active, and passive to identify the best spiral shape for enhancing the performance of the classifier. The results of the examination are compared with the results of MFO [44], PSO [46], ALO [47], SFS [48], HBO [49], and ChOA [50] algorithms for validation purposes. Table 2 presents the required primary parameters and values for each of the algorithms. Figs.9 to 14 show the convergence curves and MSE boxplot after running the algorithms 20 times for each sonar datasets. MATLAB on a PC with 16GB RAM and 3.8 GHz is used to run the algorithms. The results show that no compared algorithms can continue after 500 iterations; tables 3 to 5 presents the statistical results of the examinations,

Table 2: The initialization for comparison algorithms.

| Algorithms | Parameters | Value |
|---|---|---|
| **PSO** | $C_1, C_2$ | 2 |
| | Population size | 50 |
| **ALO** | w | [2,6] |
| | k | 500 |
| **ChOA** | m | Chaotic |
| | $r_1, r_2$ | Random [0,1] |
| | Number of chimps | 50 |
| **HBO** | $C; p_1; p_2$ | from corresponding equations [49] |
| | N | 50 |
| **SFS** | Maximum Diffusion Number (MDN) | 1 |
| **MFO** | Number of moths | 50 |

Table 3 presents the modified MFO naming for comparison purposes. The Average (AVE) ± Standard Deviation (STD) values must be lower to avoid local minima and achieve a better capability. Tables 4 to 9 show each algorithm's calculated AVE and STD; the bolded results indicate the best results achieved. The authors of [51] state the importance of statistical tests, including AVE ± STD, in a balanced performance evaluation of optimization algorithms. The tables also show the significant statistical values ($P \leq 0.05$) computed per Wilcoxon's rank-sum test [54] for each benchmark algorithm and its achieved classification rates. A significant $P$ value provides a statistical indication of whether to accept or reject a null hypothesis; for a hypothesis to be accepted, its computed $P$ value should be $P \leq 0.05$; otherwise, it is rejected. In the tables, N/A indicates that the determined algorithm cannot be compared with itself (i.e., Not Applicable).

Table 3: The naming style for the modification of MFO

| Spiral types | Archimedean | Logarithmic | Fermat | Lituus | Equiangular | Random |
|---|---|---|---|---|---|---|
| Name | LMFO1 | LMFO2 | LMFO3 | LMFO4 | LMFO5 | LMFO6 |

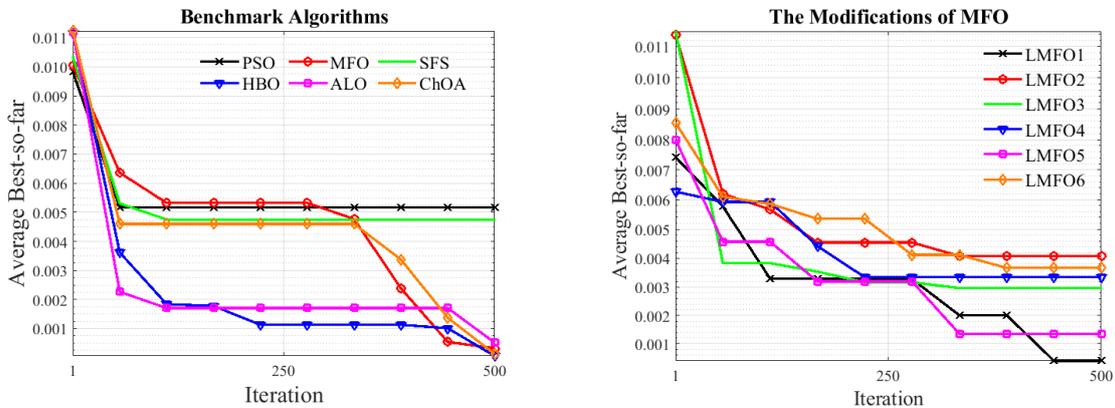

Fig.9: Convergence curves for Sejnowski & Gorman dataset.

Table 4: Results for Sejnowski & Gorman dataset (Benchmark Algorithm).

| Algorithms | PSO | ALO | ChOA | HBO | SFS | MFO |
|---|---|---|---|---|---|---|
| MSE (AVE) | 0.0051 | 0.00051 | **0.00062** | 0.00096 | 0.0051 | 0.00045 |
| MSE (STD) | 0.000075 | 0.000021 | **0.0000002** | 0.000052 | 0.00014 | 0.000052 |
| P-values | 9.28E-18 | 1.33E-12 | N/A | 0.00021 | 0.00052 | **0.065** |
| Classification rate % | 93.0214 | 94.5412 | **97.2548** | 96.3254 | 93.2177 | 96.9852 |

Table 5: Results of the modifications of MFO for Sejnowski & Gorman dataset.

| Algorithms | LMFO1 | LMFO2 | LMFO3 | LMFO4 | LMFO5 | LMFO6 |
|---|---|---|---|---|---|---|
| MSE (AVE) | **0.00058** | 0.0048 | 0.0031 | 0.0033 | 0.0018 | 0.0044 |
| MSE (STD) | **3.73E-11** | 0.0011 | 0.000035 | 0.000045 | 0.00022 | 0.0009 |
| P-values | **N/A** | 1.44E-09 | 0.00052 | 0.00019 | **0.066** | 0.0027 |
| Classification rate % | **98.8527** | 91.5018 | 95.2005 | 95.0011 | 97.2177 | 92.3344 |

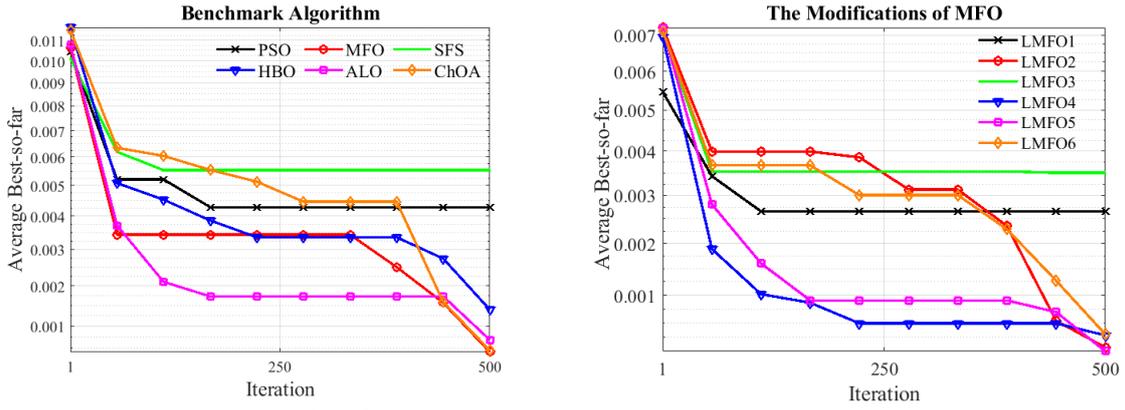

**Fig.10:** Convergence curves for passive dataset.

Table 6: Results for passive dataset (Benchmark Algorithm).

| Algorithms | PSO | ALO | ChOA | HBO | SFS | MFO |
|---|---|---|---|---|---|---|
| **MSE (AVE)** | 0.0041 | 0.00079 | **0.00073** | 0.0012 | 0.0052 | 0.00074 |
| **MSE (STD)** | 1.75E-07 | 2.01E-08 | **1.47E-07** | 0.00033 | 0.00013 | 0.000001 |
| **P-values** | 0.000007 | 0.00031 | N/A | 0.00072 | 0.00048 | **0.122** |
| **Classification rate %** | 92.4414 | 97.8520 | **98.0014** | 96.5237 | 91.9999 | 98.0009 |

Table 7: Results of the modifications of MFO for passive dataset.

| Algorithms | LMFO1 | LMFO2 | LMFO3 | LMFO4 | LMFO5 | LMFO6 |
|---|---|---|---|---|---|---|
| **MSE (AVE)** | 0.0028 | 0.00046 | 0.0034 | 0.000055 | **0.00045** | 0.00058 |
| **MSE (STD)** | 0.0004 | 1.61E-09 | 0.000023 | 0.000005 | **1.51E-09** | 1.12E-07 |
| **P-values** | 0.00024 | 4.22E-07 | 0.00042 | 0.00033 | **N/A** | 3.24E-06 |
| **Classification rate %** | 96.3324 | 98.6302 | 95.9865 | 98.0321 | **98.9998** | 98.0044 |

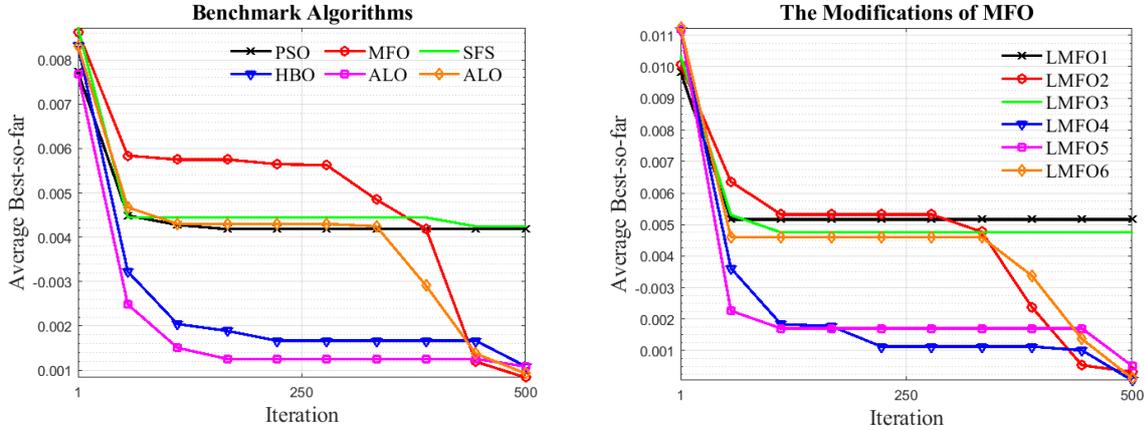

**Fig.11:** Comparison of convergence curve for active dataset.

Table 8: Results for active dataset (Benchmark Algorithm).

| Algorithms | PSO | ALO | ChOA | HBO | SFS | MFO |
|---|---|---|---|---|---|---|
| **MSE (AVE)** | 0.0042 | 0.00110 | 0.00092 | 0.00112 | 0.0045 | **0.0009** |
| **MSE (STD)** | 0.000037 | 0.00075 | 0.00071 | 0.00073 | 0.00029 | **0.00033** |
| **P-values** | 0.0054 | 1.82E-08 | **0.081** | 0.00002 | 0.00085 | N/A |
| **Classification rate %** | 92.9004 | 94.5023 | 95.9743 | 94.5284 | 90.0028 | **96.0002** |

Table 9: Results of the modifications of MFO for active dataset.

| Algorithms | LMFO1 | LMFO2 | LMFO3 | LMFO4 | LMFO5 | LMFO6 |
|---|---|---|---|---|---|---|
| **MSE (AVE)** | 0.0050 | 0.0009 | 0.0048 | **0.0008** | 0.0018 | 0.0009 |
| **MSE (STD)** | 0.00012 | 0.00002 | 0.000022 | **2.66E-12** | 0.00022 | 0.000001 |
| **P-values** | 0.022 | **0.081** | 0.0023 | N/A | **0.066** | **0.085** |
| **Classification rate %** | 94.9937 | 97.4448 | 95.2117 | **98.0881** | 97.2177 | 98.0004 |

Results presented in Tables 4, 5, and Fig.9 show that among the benchmark algorithms, LMFO has achieved the best results among comparison algorithms, and also among the modified algorithms LMFO1 and LMFO5 have achieved better results for the Sejnowski & Gorman dataset. As for the best spiral shape, the results show that Equiangular and Archimedean spiral can enhance the performance of the MFO-DCNN. The passive dataset results presented in Table 6 and 7 and Fig.10 leads us to similar conclusions as MFO, and Equiangular spiral shapes have achieved the best results. The Equiangular and Archimedean shapes presented in Fig.7 can converge faster than the rest, and they have a distinctive exploitation ability; these are why these shapes have a more significant improvement impact on the performance of MFO. The convergence curves of the algorithms are shown in Fig. 9- 11, and the graphs show that Equiangular and Lituus spiral shapes have the best convergence rates among all shapes for the selected datasets in descending order.

Contrary to the two previous datasets, MFO and LMFO4 provide the best results on benchmark algorithms and improved versions of MFO for the active sonar dataset. These results were predictable because the active sonar contains inhomogeneous phenomena such as false alarm rate and clutter; therefore, this dataset has a different nature; thereby, different algorithms present better results than other datasets.

Finally, from the results presented, it can be noted that the Equiangular spiral shape can converge better and boost the search performance of the MFO, especially in the exploitation phase. This means the searching capability of the algorithm is not easily overcome by local minima on the one hand and reaches global minima in less time on the other.

### 4.1. The Analysis of Time Complexity

Time complexity measurement is an essential step in analyzing classifiers in real-time in the sonar signal processing domain. In order to measure the time complexity of the MFO-DCNN, it is implemented and used to test a dataset containing 3364 images. The implementation was carried out in a device that had Tesla K20 as the GPU and an Intel Core i7-7700HQ up to a 3.8 GHz processor as the CPU. The result of the test is presented in Table 10. The results show that the modified DCNNs outperform the classic DCNN algorithms concerning training and testing time. More precisely, considering that 90% of the required processing time is spent on feature extraction, the DCNNs take less than a millisecond to process an image.

**Table 10:** The training and test time of comparison models implemented on CPU and GPU.

| Model | CPU vs. GPU | Training time | Testing time | P-value |
|---|---|---|---|---|
| DCNN | GPU | **342 sec** | **2225 ms** | N/A |
|  | CPU | **304 min** | **3.5 min** | N/A |
| DCNN-MFO | GPU | 355 sec | 2936 ms | 1.11E-06 |
|  | CPU | 325 min | 4.23 min | 1.22E-04 |
| DCNN-LMFOs | GPU | 354 sec | 2932 ms | 1.33E-05 |
|  | CPU | 325 min | 4.22 min | 1.71E-06 |
| DCNN-ALO | GPU | 401 sec | 3002 ms | 1.08E-07 |
|  | CPU | 333 min | 4.71 min | 1.84E-04 |
| DCNN-PSO | GPU | 359 sec | 2999 ms | 0.002 |
|  | CPU | 327 min | 4.31 min | 0.005 |
| DCNN-HBO | GPU | 425 sec | 3652 ms | 2.18E-03 |
|  | CPU | 364 min | 5.02 min | 3.56E-09 |
| DCNN-SFS | GPU | 411 sec | 3014 ms | 1.04E-07 |
|  | CPU | 341 min | 4.33 min | 1.38E-05 |
| DCNN-ChOA | GPU | 358 sec | 2998 ms | 1.84E-04 |
|  | CPU | 327 min | 4.21 min | 1.73E-06 |

### 4.2. Sensitivity Analysis of Designed Model

Here, we analyze the control parameters of the proposed classifier model, which are the shape of the logarithmic spiral represented by $q$, the number of layers represented by $N_{layer}$, and the number of batches represented by $N_{batch}$; the last two parameters belong to the network structure.

The sensitivity analysis aims to investigate the sensitivity and robustness of each parameter for various inputs. Following a similar direction as [52], four-parameter levels were used to conduct several experiments to generate an orthogonal array to characterize various parameter combinations. Table 11 shows the four-parameter levels used, and Table 12 shows the results of various parameter combinations and calculated MSEs obtained during the experiments. Also, Fig. 12 shows the parameter level trends. To summarize the results, the parameter combinations $N_{Layer} = 5$, $q = 1$, and $N_{batch} = 10$ achieves the best performance.

**Table 11:** the specification of parameters

| Level | $N_{layer}$ | $q$ | $N_{batch}$ |
|---|---|---|---|
| 1 | 3 | 0.20 | 6 |
| 2 | 4 | 0.40 | 8 |
| 3 | 5 | 0.80 | 10 |
| 4 | 6 | 1 | 12 |

**Table 12:** Results of various parameter combinations

| Experiment number | Parameters | | | Result (MSE) |
|---|---|---|---|---|
| | $N_{layer}$ | $q$ | $N_{batch}$ | |
| *Number* 1 | 1 | 0.20 | 1 | 0.0573 |
| *Number* 2 | 1 | 0.40 | 2 | 0.0451 |
| *Number* 3 | 1 | 0.80 | 3 | 0.0255 |
| *Number* 4 | 1 | 1 | 4 | 0.0142 |
| *Number* 5 | 2 | 0.20 | 2 | 0.0421 |
| *Number* 6 | 2 | 0.40 | 1 | 0.0352 |
| *Number* 7 | 2 | 0.80 | 4 | 0.0123 |
| *Number* 8 | 2 | 1 | 3 | 0.0092 |
| *Number* 9 | 3 | 0.20 | 1 | 0.0213 |
| *Number* 10 | 3 | 0.40 | 4 | 0.0101 |
| *Number* 11 | 3 | 0.80 | 2 | 0.0052 |
| *Number* 12 | 3 | 1 | 3 | 0.0013 |
| *Number* 13 | 4 | 0.20 | 4 | 0.0267 |
| *Number* 14 | 4 | 0.40 | 3 | 0.0117 |
| *Number* 15 | 4 | 0.80 | 2 | 0.0059 |
| *Number* 16 | 4 | 1 | 1 | 0.0014 |

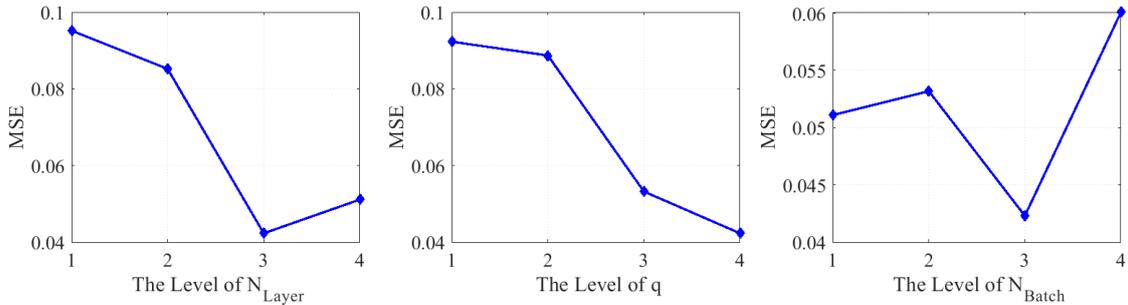

**Fig. 12:** Level trends of the analyzed parameters

## 5. Conclusions

In order to recognize different underwater sonar datasets, this chapter proposed using the Moth Flame Optimization (MFO) algorithm for fine-tuning a Deep Neural Network. Besides, this chapter investigated the efficiency of six spiral motions with different curvatures and slopes in the performance of the MFO (i.e., LMFO) for underwater sonar dataset classification tasks. In order to evaluate the performance of the customized model, in addition to benchmark Sejnowski & Gorman's dataset, two experimental sonar datasets, i.e., the passive sonar dataset, the active dataset, were exploited. The results of MFO and its modifications were compared to four novel nature-inspired algorithms, including HBO, ChOA, ALO, SFS, and the classic PSO. The results confirmed that the LMFO shows better performance than the other state-of-the-art models so that the classification rates were increased by 1.5979, 0.9985, and 2.0879 for Sejnowski & Gorman, passive, and active datasets, respectively. The results also approved that time complexity is not significantly increased by using different spiral motions.

For future research direction, it would be probable to employ LMFO in solving real-world engineering problems. Furthermore, another future research might use other spiral forms with a wider range of curvatures, slopes, and interception spots. Moreover, it can be attractive to mathematically calculate the best spiral motion to the best DCNN's performance.

**References**

[1] Ravakhah S, Khishe M, Aghababaie M, Hashemzadeh E. Sonar false alarm rate suppression using classification methods based on interior search algorithm. IJCSNS 2017;17:58.
[2] Qiao W, Khishe M, Ravakhah S. Underwater targets classification using local wavelet acoustic pattern and Multi-Layer Perceptron neural network optimized by modified Whale Optimization Algorithm. Ocean Eng 2021;219:108415. https://doi.org/https://doi.org/10.1016/j.oceaneng.2020.108415.
[3] Khishe M, Mosavi MR. Improved whale trainer for sonar datasets classification using neural network. Appl Acoust 2019. https://doi.org/10.1016/j.apacoust.2019.05.006.
[4] Pearce SK, Bird JS. Sharpening sidescan sonar images for shallow-water target and habitat classification with a vertically stacked array. IEEE J Ocean Eng 2013;38:455–69.
[5] Mosavi MR, Khishe M, Hatam Khani Y, Shabani M. Training radial basis function neural network using stochastic fractal search algorithm to classify sonar dataset. Iran J Electr Electron Eng 2017. https://doi.org/10.22068/IJEEE.13.1.10.
[6] Fialkowski JM, Gauss RC. Methods for identifying and controlling sonar clutter. IEEE J Ocean Eng 2010;35:330–54.
[7] Fandos R, Zoubir AM, Siantidis K. Unified design of a feature-based ADAC system for mine hunting using synthetic aperture sonar. IEEE Trans Geosci Remote Sens 2013;52:2413–26.
[8] Lyon RF. Human and Machine Hearing. 2017. https://doi.org/10.1017/9781139051699.
[9] Demertzis K, Iliadis L, Anezakis V-D. A deep spiking machine-hearing system for the case of invasive fish species. 2017 IEEE Int. Conf. Innov. Intell. Syst. Appl., IEEE; 2017, p. 23–8.
[10] Guo Y, Liu Y, Oerlemans A, Lao S, Wu S, Lew MS. Deep learning for visual understanding: A review. Neurocomputing 2016. https://doi.org/10.1016/j.neucom.2015.09.116.
[11] Goodfellow I, Bengio Y, Courville A, Bengio Y. Deep learning. vol. 1. MIT press Cambridge; 2016.
[12] Ruder S. An overview of gradient descent optimization algorithms. ArXiv Prepr ArXiv160904747 2016.
[13] Shewchuk JR. An introduction to the conjugate gradient method without the agonizing pain 1994.
[14] Martens J. Deep learning via Hessian-free optimization. ICML 2010 - Proceedings, 27th Int. Conf. Mach. Learn., 2010.
[15] Vinyals O, Povey D. Krylov subspace descent for deep learning. J. Mach. Learn. Res., 2012.
[16] Hinton GE, Salakhutdinov RR. Reducing the dimensionality of data with neural networks. Science (80- ) 2006. https://doi.org/10.1126/science.1127647.
[17] Abdel-Basset M, Abdel-Fatah L, Sangaiah AK. Metaheuristic algorithms: A comprehensive review. Comput Intell Multimed Big Data Cloud with Eng Appl 2018:185–231.
[18] Kozek T, Roska T, Chua LO. Genetic algorithm for CNN template learning. IEEE Trans Circuits Syst I Fundam Theory Appl 1993;40:392–402.
[19] Rosa G, Papa J, Marana A, Scheirer W, Cox D. Fine-tuning convolutional neural networks using Harmony Search. Lect. Notes Comput. Sci. (including Subser. Lect. Notes Artif. Intell. Lect. Notes Bioinformatics), 2015. https://doi.org/10.1007/978-3-319-25751-8_82.
[20] Fedorovici L-O, Precup R-E, Dragan F, Purcaru C. Evolutionary optimization-based training of convolutional neural networks for OCR applications. 2013 17th Int. Conf. Syst. Theory, Control Comput., IEEE; 2013, p. 207–12.
[21] Wang Y, Zhang H, Zhang G. cPSO-CNN: An Efficient PSO-Based Algorithm for Fine-Tuning Hyper-Parameters of Convolutional Neural Networks. Swarm and Evolutionary Computation 49: 114–123 2019.
[22] Guo B, Hu J, Wu W, Peng Q, Wu F. The tabu_genetic algorithm: a novel method for hyper-parameter optimization of learning algorithms. Electronics 2019;8:579.
[23] Kumaran N, Vadivel A, Kumar SS. Recognition of human actions using CNN-GWO: a novel

modeling of CNN for enhancement of classification performance. Multimed Tools Appl 2018;77:23115–47.
[24] Rere LM, Fanany MI, Arymurthy AM. Metaheuristic algorithms for convolution neural network. Comput Intell Neurosci 2016;2016.
[25] Ayumi V, Rere LMR, Fanany MI, Arymurthy AM. Optimization of convolutional neural network using microcanonical annealing algorithm. 2016 Int. Conf. Adv. Comput. Sci. Inf. Syst., IEEE; 2016, p. 506–11.
[26] Mohapatra P, Chakravarty S, Dash PK. An improved cuckoo search based extreme learning machine for medical data classification. Swarm Evol Comput 2015;24:25–49.
[27] Li L-L, Sun J, Tseng M-L, Li Z-G. Extreme learning machine optimized by whale optimization algorithm using insulated gate bipolar transistor module aging degree evaluation. Expert Syst Appl 2019;127:58–67.
[28] Webb GI, Keogh E, Miikkulainen R, Miikkulainen R, Sebag M. No-Free-Lunch Theorem. Encycl. Mach. Learn., 2011. https://doi.org/10.1007/978-0-387-30164-8_592.
[29] Yamany W, Fawzy M, Tharwat A, Hassanien AE. Moth-flame optimization for training multi-layer perceptrons. 2015 11th Int. Comput. Eng. Conf., IEEE; 2015, p. 267–72.
[30] Faris H, Aljarah I, Mirjalili S. Evolving Radial Basis Function Networks Using Moth-Flame Optimizer. Handb. Neural Comput., 2017. https://doi.org/10.1016/B978-0-12-811318-9.00028-4.
[31] Gorman RP, Sejnowski TJ. Analysis of hidden units in a layered network trained to classify sonar targets. Neural Networks 1988. https://doi.org/10.1016/0893-6080(88)90023-8.
[32] Khishe M, Mohammadi H. Passive sonar target classification using multi-layer perceptron trained by salp swarm algorithm. Ocean Eng 2019. https://doi.org/10.1016/j.oceaneng.2019.04.013.
[33] Mosavi MR, Kaveh M, Khishe M. Sonar data set classification using MLP neural network trained by non-linear migration rates BBO. fourth Iran. Conf. Eng. Electromagn. (ICEEM 2016), 2016, p. 1–5.
[34] Patterson J, Gibson A. Deep learning: A practitioner's approach. " O'Reilly Media, Inc."; 2017.
[35] Fukushima K, Miyake S. Neocognitron: A self-organizing neural network model for a mechanism of visual pattern recognition. Compet. Coop. neural nets, Springer; 1982, p. 267–85.
[36] Rumelhart DE, Hinton GE, Williams RJ. Learning representations by back-propagating errors. Nature 1986;323:533–6.
[37] Kramer MA. Nonlinear principal component analysis using autoassociative neural networks. AIChE J 1991;37:233–43.
[38] Smolensky P. Information processing in dynamical systems: Foundations of harmony theory. Colorado Univ at Boulder Dept of Computer Science; 1986.
[39] Goodfellow I, Pouget-Abadie J, Mirza M, Xu B, Warde-Farley D, Ozair S, et al. Generative adversarial nets In: Advances in Neural Information Processing Systems (NIPS) 2014.
[40] Hinton GE, Osindero S, Teh Y-W. A fast learning algorithm for deep belief nets. Neural Comput 2006;18:1527–54.
[41] Hochreiter S, Schmidhuber J. Long short-term memory. Neural Comput 1997;9:1735–80.
[42] LeCun Y. LeNet-5, convolutional neural networks. URL Http//Yann Lecun Com/Exdb/Lenet 2015;20:14.
[43] Mosavi MR, Khishe M, Akbarisani M. Neural Network Trained by Biogeography-Based Optimizer with Chaos for Sonar Data Set Classification. Wirel Pers Commun 2017. https://doi.org/10.1007/s11277-017-4110-x.
[44] Mirjalili S. Moth-flame optimization algorithm: A novel nature-inspired heuristic paradigm. Knowledge-Based Syst 2015. https://doi.org/10.1016/j.knosys.2015.07.006.
[45] Cundy HM, Lockwood EH. A Book of Curves. Math Gaz 1963. https://doi.org/10.2307/3612643.
[46] Kennedy J, Eberhart R. Particle swarm optimization. Proc. ICNN'95-international Conf. neural networks, vol. 4, IEEE; 1995, p. 1942–8.
[47] Mirjalili S. The ant lion optimizer. Adv Eng Softw 2015. https://doi.org/10.1016/j.advengsoft.2015.01.010.
[48] Salimi H. Stochastic Fractal Search: A powerful metaheuristic algorithm. Knowledge-Based

Syst 2015. https://doi.org/10.1016/j.knosys.2014.07.025.
[49]  Askari Q, Saeed M, Younas I. Heap-based optimizer inspired by corporate rank hierarchy for global optimization. Expert Syst Appl 2020;161:113702.
[50]  Khishe M, Mosavi MR. Chimp optimization algorithm. Expert Syst Appl 2020. https://doi.org/10.1016/j.eswa.2020.113338.
[51]  Derrac J, García S, Molina D, Herrera F. A practical tutorial on the use of nonparametric statistical tests as a methodology for comparing evolutionary and swarm intelligence algorithms. Swarm Evol Comput 2011. https://doi.org/10.1016/j.swevo.2011.02.002.
[52]  Wu C, Khishe M, Mohammadi M, Karim SHT, Rashid TA. Evolving deep convolutional neutral network by hybrid sine–cosine and extreme learning machine for real-time COVID19 diagnosis from X-ray images. Soft Comput 2021:1–20.